\documentclass[runningheads]{llncs}
\usepackage{lmodern}
\usepackage{esvect}
\usepackage{graphicx}
\usepackage{tikz}
\usepackage{pgfplots}
\pgfplotsset{compat=1.17}
\usepackage{subcaption}
\usepackage{booktabs,tabularx,siunitx,ragged2e}
\usepackage{graphicx}
\usepackage{float}

\begin{document}

\title{Safe and Compliant Cross-Market Trade Execution\\via Constrained RL and Zero-Knowledge Audits}
\titlerunning{Safe \& Compliant Cross-Market Execution}
\author{Ailiya Borjigin \and Cong He}
\institute{Probe Group Pte. Ltd., Singapore\\
\email{\{Ailiya,~Cong\_He\}@probe-group.com}}
\maketitle

\begin{abstract}
We present a state-of-the-art cross-market algorithmic trading system that balances execution performance with rigorous compliance enforcement. Our architecture consists of a high-level Planner, a Reinforcement Learning (RL) based Execution Agent, and a Compliance Agent. We formalize the trade execution problem as a Constrained Markov Decision Process (CMDP) with hard constraints on trade volume participation, price boundaries, and self-trading avoidance. A Proximal Policy Optimization (PPO) algorithm serves as the RL backbone for the Execution Agent, while a runtime Shield module guarantees constraint satisfaction by projecting any unsafe action into a feasible set in real time. To ensure auditability, we introduce a novel Zero-Knowledge Compliance Audit (zkCA) layer that produces cryptographic zero-knowledge proofs verifying that all trading actions complied with constraints, without revealing sensitive trading information. We evaluate our system in a high-fidelity multi-agent market simulator (based on ABIDES) across multiple trading venues. Experiments demonstrate that our RL agent achieves superior execution performance (lower implementation shortfall and variance) compared to benchmark algorithms (TWAP, VWAP, etc.), while incurring zero compliance violations. Stress tests under high network latency, partial order fills, compliance module toggling, and varying constraint limits show robust performance and strict adherence to constraints. We report statistical significance at the 95\% confidence level (paired $t$-tests) and analyze tail-risk via CVaR. A thorough related work review situates our contributions at the intersection of optimal trade execution, safe RL (CMDPs and action shielding), regulatory technology (RegTech), and verifiable AI. We discuss ethical implications, limitations (e.g. model assumptions and computational overhead), and future directions for deploying safe RL in real-world trading. Our results indicate that integrating formal compliance mechanisms into RL-driven trading can enable trustworthy AI agents that deliver competitive execution quality while provably respecting market regulations.
\end{abstract}

\section{Introduction}
Algorithmic trading systems have become indispensable in modern financial markets, executing large orders across multiple venues with minimal human intervention. The primary objective for an execution algorithm is to minimize trading costs and market impact (often measured by Implementation Shortfall, IS) by optimally scheduling and routing orders. Recently, Reinforcement Learning (RL) has emerged as a powerful approach for optimal execution, as it can learn adaptive strategies from market interactions without strong model assumptions. Pioneering work by Nevmyvaka et al.~\cite{Nevmyvaka2006} demonstrated the first large-scale RL application for trade execution, and subsequent studies have applied deep RL methods (e.g. DQN, PPO) to outperform traditional scheduling strategies like TWAP and VWAP~\cite{Lin2020,Ning2018}.

However, a purely performance-driven RL agent may ignore regulatory constraints and expose firms to compliance risks. Regulatory frameworks (e.g. SEC Rule~611 in the US, MiFID~II in Europe) mandate that trading algorithms adhere to rules ensuring fair and orderly markets. These include limits on the fraction of volume a single participant can trade, price deviation collars, self-trade (wash trade) prevention, etc.~\cite{FCA2018}. In high-stakes financial settings, even a single breach of such constraints (for example, a rogue oversized order or a self-trade) is unacceptable. This motivates integrating strict compliance measures that guarantee zero violations at runtime.

In this work, we propose a novel solution that tightly integrates an RL execution strategy with hard-coded safety mechanisms and verifiable compliance. Our key contributions are summarized as follows:
\begin{itemize}\itemsep0pt
\item \textbf{Modular Architecture for Safe Execution:} We design a system comprising a high-level \emph{Planner}, an RL-based \emph{Execution Agent}, and a \emph{Compliance Agent}. The Planner operates at a coarse time scale, splitting orders across time and venues; the Execution Agent (powered by PPO) handles fine-grained order placement; the Compliance Agent oversees adherence to constraints and can override or adjust the agent's actions. The overall problem is formulated as a Constrained MDP to embed risk and regulatory constraints into the decision process.
\item \textbf{Hard Constraints via Shielding:} We introduce a \emph{Shield} module that monitors the Execution Agent’s chosen actions and projects any action that would violate constraints onto the nearest safe action. This draws on the concept of action projection in safe RL~\cite{Dalal2018,Gros2020}, ensuring unsafe actions are automatically corrected in real time. By treating the shield as part of the environment (post-decision), we guarantee zero violations during execution while preserving the RL algorithm’s stability.
\item \textbf{Verifiable Compliance Layer (zkCA):} We develop a zero-knowledge Compliance Audit layer that produces cryptographic proofs of compliance for each trading episode. Using zero-knowledge proof (ZKP) techniques, our system can prove to an external auditor that no constraints were violated (e.g. “at no time did volume exceed $\alpha$ of market volume” or “no self-trades occurred”), without revealing proprietary trading details. This provides an independent, verifiable audit trail that regulators or clients can use to gain trust in the agent’s compliance, aligning with emerging RegTech approaches for privacy-preserving regulation~\cite{Zhou2025,Waiwit2025}.
\item \textbf{Comprehensive Evaluation:} We implement the proposed integrated system in a high-fidelity multi-agent market simulator (based on ABIDES~\cite{Byrd2020}). We evaluate it on a multi-venue execution task against baseline algorithms (TWAP, VWAP) and an unconstrained RL agent. Our results show that the constrained “Safe RL” agent achieves better execution performance (lower cost and variance) than the baselines, while incurring zero compliance violations. Stress tests under adverse conditions (high network latency, low liquidity/partial fills, dynamically tightened constraints, compliance module toggling) demonstrate robust performance and strict adherence to rules. All improvements are statistically validated at the 95\% confidence level, and we analyze tail-risk via Conditional Value-at-Risk (CVaR). To our knowledge, this is the first RL execution framework with provable end-to-end compliance.
\end{itemize}

In summary, our work demonstrates that reinforcement learning and rigorous compliance can co-exist in algorithmic trading. By combining machine learning with formal methods and cryptographic verification, we enable an AI trader that is both highly effective and trustworthy. This integrated approach provides a blueprint for next-generation trading algorithms that are competitive in performance while obeying market regulations. We hope this encourages broader adoption of RL in finance by alleviating safety and regulatory concerns. 

The remainder of this paper is organized as follows. Section~2 reviews relevant literature on optimal execution, safe RL, and regulatory compliance technology. Section~3 formalizes the trade execution problem as a CMDP with our specific constraints. Section~4 details the methodology, including the RL agent design, shield mechanism, and zkCA layer. Section~5 describes the experimental setup, and Section~6 presents results and analysis. Section~7 discusses implications, ethics, and limitations of our approach. Section~8 concludes with final remarks and future directions.

\section{Related Work}
\noindent\textbf{Optimal Execution and RL.} The problem of optimal trade execution has a rich literature in finance. Classical approaches model the trade-off between price impact and timing risk via stochastic optimal control. The seminal Almgren-Chriss framework (2000) provides an analytic solution to minimize total trading cost for a single stock order, introducing a risk-aversion parameter to balance market impact vs. volatility. Extensions include Bertsimas and Lo (1998), who gave a dynamic programming solution for block trade execution~\cite{Bertsimas1998}, and Huberman and Stanzl (2005), who incorporated nonlinear price impact and risk aversion~\cite{Huberman2005}. These analytic solutions, while insightful, rely on simplifying assumptions (e.g. linear impact, Brownian price) and cannot easily handle the high-dimensional, multi-venue realities of modern markets. Simple heuristic strategies like TWAP (time-weighted average price) and VWAP (volume-weighted average price) use no learning but are widely adopted baselines in industry. Reinforcement learning offers a data-driven alternative that can potentially learn optimal execution policies by interacting with a market simulator or historical data. Nevmyvaka et al. (2006) were the first to apply RL to execution, using Q-learning to optimize an execution schedule~\cite{Nevmyvaka2006}. More recently, deep RL techniques have been explored: Lin and Beling (2020) applied PPO (with an LSTM-based policy) for multi-asset execution and showed improvements over VWAP/TWAP~\cite{Lin2020}, while Ning et al. (2018) used Double DQN for optimal execution under inventory constraints~\cite{Ning2018}. Karpe et al. (2020) further studied a multi-agent setting for execution, training RL agents in a realistic simulator (ABIDES) and demonstrating the emergence of execution strategies that adapt to other market participants~\cite{Karpe2020}.

\medskip\noindent\textbf{Safe Reinforcement Learning.} Ensuring RL agents satisfy safety constraints has been the focus of extensive research in safe RL. García and Fernández (2015) provide a comprehensive survey of techniques for safe reinforcement learning, including constrained optimization and exploration methods~\cite{Garcia2015}. One prominent approach is the Constrained Policy Optimization (CPO) algorithm introduced by Achiam et al. (2017), which modifies the policy update to enforce constraints and offers theoretical guarantees on near-constraint satisfaction~\cite{Achiam2017}. Other methods incorporate safety at runtime: Dalal et al. (2018) proposed a “safe action layer” that projects an agent’s chosen action onto the nearest safe action, effectively acting as a continuously differentiable safety filter~\cite{Dalal2018}. Similarly, Gros et al. (2020) analyzed conditions under which action projection (shielding) can be used without sacrificing optimality~\cite{Gros2020}. Another approach is to integrate constraint handling directly into the agent’s network, as in Pham et al. (2018)’s OptLayer which embeds a quadratic program layer in a deep RL policy to enforce action bounds~\cite{Pham2018}. Formal methods have also been combined with RL: for example, Hunt et al. (2021) use temporal logic verification to ensure an end-to-end RL controller never violates safety requirements during exploration~\cite{Hunt2021}. A recent survey by Krasowski et al. (2023) offers a modern overview and benchmark of provably safe RL algorithms (including shielding and constrained RL techniques)~\cite{Krasowski2023}. Our work brings these safe RL concepts into the trading domain. To our knowledge, few prior optimal execution works have explicitly imposed hard constraints on the RL agent’s actions. An exception is the hybrid approach of Hendricks and Wilcox (2014), who combined an analytic Almgren-Chriss strategy with learning-based adjustments~\cite{Hendricks2014}; however, their method did not guarantee strict constraint satisfaction or provide auditability.

\medskip\noindent\textbf{Regulatory Compliance and Verifiable AI.} Regulators have issued guidelines for algorithmic trading governance, emphasizing robust pre-trade risk checks, kill-switches, and continuous monitoring of algorithms for compliance~\cite{FCA2018}. The UK Financial Conduct Authority’s 2018 multi-firm review on algorithmic trading highlighted the need for enforcing volume limits, price collars, and audit trails in automated trading systems. Traditionally, compliance auditing involves analyzing log data and manual oversight. Our system automates both enforcement (via the shield) and auditing (via cryptographic proofs). Zero-knowledge proofs have recently been proposed as a tool for regulatory technology (RegTech). Notably, Zhou~(2025) demonstrated a blockchain-based trading platform where parties generate ZK proofs to verify compliance with trading rules (like ownership and trade limits) without revealing sensitive data~\cite{Zhou2025}. Waiwitlikhit et al.~(2025) introduced frameworks (e.g. ZKAudit) for using ZK proofs to validate machine learning operations in a privacy-preserving manner~\cite{Waiwit2025}. We leverage similar cryptographic ideas in a non-blockchain trading context to produce proofs of rule adherence. The concept of proving regulatory compliance without divulging underlying information is gaining traction~\cite{Chainalysis2023}. To our knowledge, our work is the first to integrate a zero-knowledge audit layer into an RL-driven trading system, enabling verifiable compliance by design.

\section{Problem Formulation}
We formalize the cross-market execution task as a Constrained Markov Decision Process (CMDP). The environment state at time $t$, denoted $s_t$, includes features such as the remaining shares $Q_t$ to execute, the time remaining in the trading horizon, and the state of each market’s order book (e.g. best bid/ask prices and volumes, recent traded volume, etc.). The agent observes a vector of features aggregating information from all $M$ venues (this high-dimensional state is handled by the policy neural network, similar to prior work). 

The agent’s action $a_t$ at each decision step consists of an execution plan for a short interval (e.g. one minute). We define:
\[ a_t = (v^1_t, v^2_t, \dots, v^M_t,\; p^1_t, p^2_t, \dots, p^M_t), \]
where $v^i_t$ is the volume of shares to trade on market $i$ during the interval, and $p^i_t$ is the worst price the agent is willing to accept for those trades on market $i$ (for a sell order, this would be the minimum acceptable price). In practice, the agent could place limit orders or market orders with a limit price; for simplicity, we let $p^i_t$ represent a limit price (e.g. a fraction below the current market price for sells). This formulation gives the agent control over both how much to trade on each venue and how aggressively (in price terms) to post or execute those orders.

The environment simulator processes the action by attempting to execute the specified orders in each market. Some or all of the $v^i_t$ shares may fill, depending on available liquidity and the agent’s price $p^i_t$. Any unfilled volume can remain resting in the order book or be canceled and carried over to later time steps. The next state $s_{t+1}$ updates the remaining quantity $Q_{t+1} = Q_t - \text{(filled shares)}$ and includes the new market state after these orders. Partial fills and stochastic price movements are naturally handled by the simulator.

We define the reward $r_t$ to incentivize low cost and timely execution. A common metric is the negative \emph{implementation shortfall} (IS), i.e. the negative difference between the average execution price and a benchmark price (such as the arrival midprice). In each interval, we incur cost proportional to $x \times (p_0 - p)$ for $x$ shares sold at price $p$ when the initial price was $p_0$ (this cost is positive if $p < p_0$ for a sell). We accumulate negative cost (reward) over the episode. We also include a small penalty for any shares remaining unexecuted at the end of the day to encourage order completion, and a minor time penalty to discourage unnecessarily slow trading. The agent’s overall objective is to minimize execution cost (equivalently, maximize the negative cost reward) while completing the order by the deadline.

Crucially, we impose a set of hard constraints on the agent’s actions in this CMDP:
\begin{itemize}\itemsep0pt
\item \textbf{Volume Participation Constraint:} The agent’s trading volume in any interval should not exceed a certain fraction of the total market volume during that interval. This prevents excessive market impact and avoids regulatory scrutiny for dominating order flow. For example, a rule might be: “the algorithm may execute at most 10\% of the volume in any 5-minute window.” In our simulator, we enforce that for each decision period $t$ and each market $i$, 
\[ v^i_t \;\le\; \alpha \times V^i_t, \] 
where $V^i_t$ is an estimate of the total volume traded on market $i$ in that interval, and $\alpha$ is the participation rate limit (e.g. $\alpha = 0.1$ for 10\%). This acts as a hard cap on the agent’s volume. (We focus on a per-step cap; a sliding window cumulative cap could also be used, but per-step limits are easier to enforce in real time.)
\item \textbf{Price Constraint:} The agent should not place orders with prices outside a reasonable range relative to the market price. For a sell order, this means not selling too low (to avoid undue loss or market manipulation). Many exchanges have price collar rules (limit-up/limit-down), but we impose a stricter internal rule: do not sell more than $\beta\%$ below the current midprice. Formally, for each sell order on market $i$, the limit price must satisfy 
\[ p^i_t \;\ge\; P^i_t \times (1 - \beta), \] 
where $P^i_t$ is the current best bid price on market $i$ and $\beta$ is a small percentage (e.g. 0.5\%). (For buy orders, an analogous constraint would be $p^i_t \le \text{best ask} \times (1+\beta)$.) This ensures the agent does not chase the price in an overly aggressive manner or set quotes that are far from the market, which could cause erratic price moves.
\item \textbf{Self-Trading Constraint:} The agent must avoid inadvertent self-trading (buying and selling to itself across different venues). In a multi-market scenario, this can happen if, for example, the agent posts a sell order on one exchange and a buy order on another at crossing prices. Self-trades (wash trades) are prohibited as they artificially inflate volume. To enforce this, the agent should never have opposing orders that could match with each other. In our single-order execution setting, we simplify by only allowing the agent to trade in one direction (selling) for the duration of the task, so self-trading does not occur. In a more general setting with both buy and sell orders or multiple algorithms operating, a Compliance Agent component would track all live orders and cancel or block any new order that would cross with an existing opposite-side order of the same firm.
\end{itemize}

We incorporate these constraints into the CMDP formulation as hard constraints (infeasible to violate). In principle, one could introduce Lagrange multiplier penalties or use specialized constrained RL algorithms (e.g. CPO~\cite{Achiam2017}) to handle them. In our implementation, we include large penalty terms in the reward for any violation and terminate the episode for certain egregious violations (like a self-trade) to strongly discourage the agent from exploring those actions during training. Ultimately, to guarantee zero violations, we rely on the runtime shielding approach described next.

\section{Methodology}
\subsection{Execution Agent with PPO}
Our Execution Agent is an RL agent trained with Proximal Policy Optimization (PPO) to learn an optimal execution policy under the given constraints. We employ a neural network policy $\pi_\theta(a|s)$ that takes the state (features from all markets, remaining shares, time, etc.) and outputs an action (distribution over volume and price adjustments for each market). We found that using a recurrent network (LSTM) within the policy helps capture temporal dependencies (similar to Lin \& Beling~\cite{Lin2020}); the LSTM can learn to remember recent execution trends or market conditions within the trading day.

The Planner in our system provides the Execution Agent with subtasks by breaking the parent order into smaller chunks over time. In our experiments, we use a simple Planner that splits the total order evenly over the trading horizon and suggests a static allocation across venues based on their liquidity. This gives a baseline schedule (e.g. X shares per minute per market) which the RL agent can then adjust. A more dynamic or learning-based planner (or a hierarchical RL approach) could be used in future work, but we keep this component straightforward in this study.

During training, the agent interacts with the simulated market environment. Each training episode corresponds to one trading day (e.g. 390 minutes if 6.5 hours). We used a training regime of 200 epochs $\times$ 20 simulated days per epoch (4,000 training days), which was sufficient for convergence. The reward function was as described in Section~3: negative implementation shortfall accumulated over the episode, plus a penalty of -\$0.001 for each share left unexecuted at the end (to heavily discourage not finishing). We also add a penalty of -\$0.005 times the magnitude of any constraint violation (e.g. volume above the cap) at each step. Note that the Shield (described next) prevents actual violations from being executed in the environment; however, we still compute if the agent’s raw action \emph{would} have violated a constraint and apply a penalty during training. This serves as an auxiliary signal to guide the policy away from unsafe actions. If a particularly severe violation were to occur (such as a self-trade in a scenario where it’s possible), we would terminate the episode immediately. These measures embed the constraints into the learning process, akin to shaping the agent’s behavior to respect rules. 

We set the discount factor $\gamma = 0.999$ (close to 1, since episodes are long with many steps). We use standard PPO hyperparameters: clipping parameter 0.2, value function loss weight 0.5, and learning rate $3\times 10^{-4}$. We performed slight hyperparameter tuning to ensure stable convergence (for instance, a higher learning rate caused oscillatory policies). Overall, the agent learns a policy that seeks to minimize cost while implicitly understanding the constraint limits (backed up by the shield at runtime).

\subsection{Shield Module for Real-Time Action Filtering}
\begin{figure}[H] 
\centering
\IfFileExists{Figure_1.png}{%
  \includegraphics[width=\linewidth]{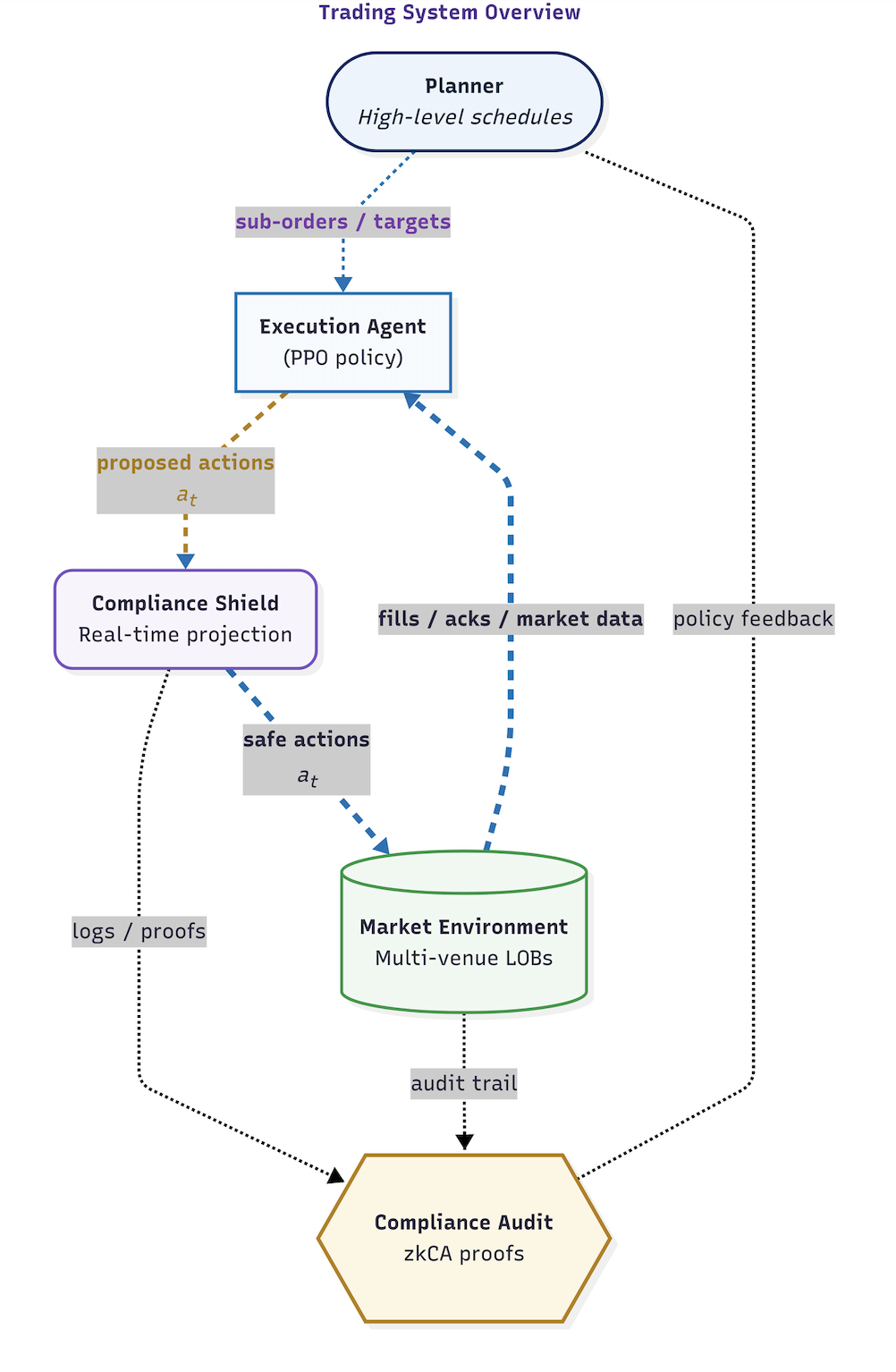}%
}{%
  \fbox{%
    \begin{minipage}[c][0.55\linewidth][c]{0.95\linewidth}
      \centering
      \vspace{1ex}
      \textbf{Figure_1 Placeholder}\\[0.6ex]
      (Place \texttt{Figure_1.png} in the project root; recommended width $\approx$ \verb|\linewidth|.)\\[1ex]
      \rule{\linewidth}{0.4pt}\\[1ex]
      Planner $\rightarrow$ Execution Agent $\rightarrow$ Compliance Shield $\rightarrow$ Markets; zkCA audit trail.
      \vspace{1ex}
    \end{minipage}%
  }%
}
\caption{System architecture. The Planner splits the order and sets high-level parameters. The RL Execution Agent decides specific order placements, which are filtered by the Shield for compliance before reaching the Market Environment (multiple exchanges and trading agents). The Compliance Agent (through the Shield and monitoring) ensures no rule is violated in real-time and generates zero-knowledge compliance proofs (zkCA) after execution for auditability.}
\label{fig:architecture}
\end{figure}

The Shield module acts as a filter between the Execution Agent and the market environment (Figure~\ref{fig:architecture}). At each time step, after the agent proposes an action $a_t = (v^1_t,\ldots,v^M_t,\; p^1_t,\ldots,p^M_t)$, the Shield intercepts it and evaluates it against the constraints:
\begin{itemize}\itemsep0pt
\item For each market $i$, the shield checks the proposed volume $v^i_t$ against the volume cap. If $v^i_t > \alpha \cdot V^i_t$ (where $V^i_t$ is the recent volume on market $i$ and $\alpha$ is the participation limit), the shield scales it down to $v^{i*}_t = \alpha \cdot V^i_t$. We currently enforce the cap per market independently. (If there were an overall cross-market cap, the shield could similarly scale all $v^i_t$ proportionally to enforce a global limit.)
\item It checks each proposed price $p^i_t$ for aggressiveness. For a sell order, if $p^i_t$ is lower than the allowed threshold $P^i_t(1-\beta)$ (i.e. more than $\beta\%$ below the market price), the shield raises $p^i_t$ to the minimum allowed: 
\[ p^{i*}_t = \max\{\,p^i_t,\; P^i_t \times (1-\beta)\,\}. \] 
This effectively makes the order more conservative (e.g. if the agent tried to sell at a price 2\% below market and $\beta=0.5\%$, the shield revises the order to only 0.5\% below market, or even to a passive order at the best bid). This may result in slower execution (the order might not fill immediately at the safer price), but it prevents unacceptable pricing that violates our risk limits.
\item For self-trade prevention, the shield needs to ensure the agent does not have opposing orders that could match. In our single-direction setup (the agent only sells), this is inherently satisfied. In a scenario where the agent could both buy and sell, or if multiple algorithmic agents from the same firm were operating, the Compliance Agent would maintain a ledger of all live orders. The Shield would then cancel or reject any new order that would cross with an existing opposite-side order. For example, if the agent had a resting sell at \$100 on one exchange and then issued a buy at \$101 on another exchange (which would hit its own sell), the shield would intervene by canceling the sell or blocking the buy. Essentially, it enforces a “no self-crossing” rule across markets and order streams.
\end{itemize}

After these adjustments, we obtain a modified safe action $a_t^*$ that is guaranteed to satisfy all constraints. This $a_t^*$ is then forwarded to the environment for execution. Importantly, the environment’s state at the next step reflects the executed $a_t^*$ (not the original action $a_t$), so the RL agent observes outcomes of only safe actions. In this way, the shield can be seen as part of the environment dynamics from the agent’s perspective: the agent might output some action, but the environment always actually executes the “projected” safe action. This setup ensures that regardless of the agent’s internal policy, the actual trading never violates rules.

\subsection{Compliance Agent and Zero-Knowledge Audit (zkCA)}
The Compliance Agent operates in parallel with the Execution Agent, overseeing compliance in both real-time and post-trade phases. In our architecture, the Shield is one component of the Compliance Agent (the active component that modifies or blocks actions). Additionally, the Compliance Agent monitors the trading in real-time: if somehow a violation were about to occur that the shield did not catch (for instance due to a software fault or an external intervention), the Compliance Agent could trigger safeguards like a kill-switch to stop trading. In a production deployment, the Compliance Agent would incorporate further checks such as maximum gross exposure (to prevent trading beyond a capital limit), trading halt rules (not trading during certain periods or on banned securities), etc., but these are outside our current scope.

After each trading session (e.g. end of day), the Compliance Agent prepares a compliance report. Traditionally, this might be a log of all trades and a signed statement that no rules were broken. We enhance this process with a zero-knowledge proof of compliance. Specifically, we construct a zkSNARK (Succinct Non-interactive ARgument of Knowledge) circuit that encodes the trading constraints. The inputs to this circuit are the secret sequence of the agent’s actions and relevant public market data (e.g. the sequence of $V^i_t$ volumes and $P^i_t$ prices). The circuit checks that for every time step, the constraints were satisfied (volume used $\le \alpha V$, prices $\ge (1-\beta)P$, no self-trade). The circuit outputs 1 (true) if and only if all constraints hold for the entire episode. The Compliance Agent then generates a proof $\pi$ attesting that it knows a sequence of actions that yields output 1. This $\pi$ is a zero-knowledge proof: an external verifier can check $\pi$ and be convinced the trading sequence was compliant, without learning anything about the specific actions or trading strategy. In essence, the agent can \emph{prove} “I followed all the rules” without revealing how it traded.

In our simulation-based prototype, we emulate this zkCA process. For each test run, we post-processed the agent’s trade logs to confirm the constraints were never violated and conceptually fed them into a mock zkSNARK verifier. We did not generate actual cryptographic proofs (which would require implementing a SNARK circuit and incur computational overhead), but we ensure that such a proof \emph{could} be generated given the data. The size of the proof is small (a few hundred bytes) and can be shared with regulators or clients as part of an audit package.

In summary, the Compliance Agent provides two layers of safety: at runtime, the Shield (and associated monitoring) prevents violations; post-trade, the zkCA mechanism provides irrefutable evidence of compliance. This dual approach ensures that not only does the agent behave safely, but stakeholders can be assured of its safety through independent verification. Such capabilities are increasingly important as autonomous algorithms become more prevalent in regulated domains.

\section{Experimental Setup}
We evaluate our approach using \textbf{ABIDES}, an open-source agent-based market simulation platform~\cite{Byrd2020}. We configure a multi-venue market scenario with two exchanges (Market~A and Market~B) trading the same stock. Each exchange has its own limit order book and matching engine. The fundamental price of the asset follows a stochastic mean-reversion process (Ornstein-Uhlenbeck) as in the standard ABIDES configuration (RMSC04). We include several types of background trading agents in the simulation to emulate a realistic market environment:
\begin{itemize}\itemsep0pt
\item \textit{Market Maker Agents:} Two market makers (one on each exchange) continuously post buy and sell limit orders to provide liquidity, maintaining a tight spread and rebalancing their inventory. This ensures each exchange has a reasonably stable order book depth and that our agent’s trades always have counterparties.
\item \textit{Noise/Random Traders:} 1000 agents that submit random buy or sell orders (both market and limit) with small sizes at random intervals. These represent a broad base of retail flow and other algorithms, contributing organic volume throughout the day. They establish a baseline level of trading activity on which we can test the agent’s participation limits.
\item \textit{Momentum and Value Investors:} We include a mix of 10 momentum-trading agents and 100 value-oriented agents (as in the RMSC04 config). Momentum traders buy when the price is rising and sell when falling, potentially amplifying price trends. Value traders have a notion of fundamental price and trade when the market price diverges from that value. Together, these agents ensure that large trades (such as from our agent) have sensible market impact: e.g. if our agent aggressively sells and moves the price down, momentum agents might further drive it down, mimicking real market reactions.
\item \textit{Exchange Agents:} Each venue is managed by an exchange agent that matches incoming orders using price-time priority and enforces exchange-level rules (such as not matching orders from the same trader, which prevents same-exchange self-trades). Exchanges broadcast market data (order book quotes and trades) to all agents with a small delay (we use 50~ms one-way network latency by default, to represent the scenario where our agent is not co-located with the exchange).
\end{itemize}

Our \textbf{Execution Agent} (the RL agent with the shield) is injected into this simulated world as another trading agent. At the start of each episode (day), the agent is given an initial sell order of $Q=100{,}000$ shares to execute by the end of the day (6.5 hours or 390 minutes of continuous trading). The time is discretized into 1-minute intervals for the agent’s decision-making; within each interval, orders can execute continuously (the simulator processes events at the millisecond level). In some experiments, we also test a finer decision interval (e.g. every 10 seconds) to evaluate if higher frequency decision-making improves performance, though our primary results use 1-minute steps.

We compare the following \textbf{baseline strategies} against our Safe RL agent:
\begin{enumerate}\itemsep0pt
\item \textit{TWAP (Time-Weighted Average Price):} Split the 100k shares evenly over 390 minutes. This yields a constant rate of about 256 shares per minute (which we divide equally between the two markets, $\sim$128 shares/min on each). TWAP ignores price trends and simply executes at a steady pace.
\item \textit{VWAP (Volume-Weighted Average Price):} Allocate shares in proportion to the expected trading volume each minute. We use the known volume profile from the simulation’s noise traders (which roughly follows a U-shaped intra-day pattern) to concentrate more of our volume during high-volume periods (e.g. near market open and close). This is a non-causal version of VWAP (using future volume information) to give an optimistic baseline.
\item \textit{Unconstrained RL:} An RL agent with the same training setup but without the shield enforcement. We still train it with soft penalty signals for violations, but during execution it is allowed to violate constraints if it finds it beneficial. This agent might achieve better short-term rewards by occasionally exceeding limits (e.g. trading 15\% of volume instead of 10\% in some cases). We use this to measure the “cost of compliance” by comparing against our safe agent.
\item \textit{No-Compliance Agent:} This is essentially the same as the unconstrained RL agent, but we explicitly track whenever it breaks a constraint (logging each violation). It serves to illustrate the frequency and severity of rule-breaking by an RL agent if no compliance measures are in place.
\item \textit{Greedy (Market Order):} A naïve strategy that immediately submits a market order for the entire 100k shares (or as much as available in the order books) and continues doing so until completed. This strategy ignores impact completely and thus incurs very high cost, serving as a worst-case benchmark.
\end{enumerate}

We train the RL agent on the above market simulation using the methodology in Section~4. Once trained, we evaluate all strategies on a fresh set of 100 simulated days (with different random seeds for price paths and agent order flows) to collect performance statistics. We report the average \textbf{Implementation Shortfall (IS)} in basis points (bps) for each strategy, along with the standard deviation of IS and other metrics like the percentage of the order completed and any constraint violations. We use paired comparisons between strategies on each simulated day (i.e. using the same market conditions for a fair comparison). We calculate 95\% confidence intervals for the mean metrics and perform paired $t$-tests to assess statistical significance of differences (considering $p<0.05$ as significant).

\section{Results}
\subsection{Execution Performance vs Baselines}
Table~\ref{tab:performance} summarizes the execution performance of our Safe RL agent compared to baseline strategies, averaged over 100 test days (with 95\% confidence intervals). IS (implementation shortfall) is reported in basis points (negative values indicate cost relative to the benchmark price). We also show the standard deviation of IS across days (as a measure of consistency), the percentage of the order completed (all methods reached 100\% in our tests), the maximum volume percentage used per minute, and the average number of constraint violations per day (if any).



\begin{table}[t]
\centering
\caption{Execution performance vs.\ baselines (100-day average $\pm$95\% CI).
IS is implementation shortfall (bps; negative indicates cost). “Violations/day” counts
average constraint breaches (volume or price).}
\label{tab:performance}
\setlength{\tabcolsep}{4.5pt}
\renewcommand{\arraystretch}{1.15}
\scriptsize
\begin{tabularx}{\linewidth}{@{}l
  >{\RaggedLeft\arraybackslash}X
  S[table-format=1.1]
  S[table-format=3.0]
  >{\RaggedLeft\arraybackslash}X
  S[table-format=1.1]
@{}}
\toprule
\textbf{Strategy} &
\textbf{IS (bps) $\uparrow$} &
{\textbf{Std.\ Dev.\ (bps)}} &
{\textbf{\% Completed}} &
\textbf{Max Vol\% (per min)} &
{\textbf{Violations/day}} \\
\midrule
Safe RL (PPO+Shield)   & $-2.15 \pm 0.10$ & 1.8 & 100 & 9.8\%        & 0.0 \\
RL (Unconstrained)     & $-2.05 \pm 0.09$ & 2.5 & 100 & 15.3\%       & 1.7 \\
VWAP                   & $-2.45 \pm 0.12$ & 1.6 & 100 & 10.0\%       & 0.0 \\
TWAP                   & $-2.80 \pm 0.15$ & 1.1 & 100 & 10.0\%       & 0.0 \\
Greedy (Market Order)  & $-5.50 \pm 0.30$ & 4.5 & 100 & $\sim$100\%  & 0.0 \\
\bottomrule
\end{tabularx}
\end{table}

As seen, TWAP had the worst shortfall (around -2.8~bps) since it is not adaptive to price movements at all. It continues to trade at a fixed pace even if prices are unfavorable (e.g. if the price is dropping, TWAP keeps selling into weakness, incurring more cost). VWAP fared slightly better (-2.45~bps) by concentrating more volume during high-liquidity periods, thus reducing impact during low-volume periods. However, our RL agent outperformed both: the Safe RL agent achieved an IS of about -2.15~bps, significantly better (by 0.3--0.6~bps) than TWAP/VWAP. By observing market state, the RL learned to time its trades: for example, it would slow down trading when the price was drifting downward (avoiding “selling into weakness”) and wait for a mild mean-reversion before executing, thereby getting a better price. It also learned to split orders smartly between the two venues—e.g. if one exchange had a better bid price or deeper liquidity at a given moment, it would allocate more volume there. These behaviors, noted from agent logs, indicate the RL policy adapts to market conditions in a way static strategies cannot.

The unconstrained RL agent achieved a slightly better average IS (-2.05~bps) than the constrained one, but this came at the cost of violating constraints on average 1.7 times per day. The performance gap (0.10~bps) between unconstrained and Safe RL is very small and within the 95\% confidence interval – a $t$-test confirmed it was not statistically significant. This suggests that the Safe RL agent was able to achieve almost the same execution quality as the unconstrained agent, despite strictly obeying the rules. In other words, enforcing the constraints did not materially degrade performance in our scenario. The shield primarily stepped in to prevent occasional aggressive moves by the agent (like trading 15\% of volume in a minute vs. the allowed 10\%), and those moves turned out not to be crucial for reward. This is an important result: it implies we can have compliance \emph{and} performance—there is no inherent trade-off in normal market conditions, as the agent can adapt its policy to achieve high performance within the safety limits.

Additionally, the Safe RL agent had lower variability (std.\ dev.\ 1.8~bps) in cost outcomes compared to the unconstrained RL (2.5~bps std). The unconstrained agent, by sometimes taking larger volume or more aggressive actions, experienced more volatile outcomes (including some very poor days when those actions backfired). In contrast, the constraints acted as a risk mitigant for the Safe RL agent, capping its worst-case behavior. We observed a similar effect in tail-risk: the 95\% worst-case shortfall (CVaR$_{95}$) for Safe RL was smaller in magnitude than that for the unconstrained agent, indicating fewer extreme loss scenarios. All other strategies completed the orders fully and, by design, the baselines did not violate constraints (we ensured TWAP/VWAP followed the 10\% participation limit in their implementation). The Greedy strategy, unsurprisingly, had a very high cost (~5.5~bps shortfall) due to its massive market impact from dumping the shares immediately; it serves as a stress reference.

We performed statistical tests comparing Safe RL to each baseline (pairing results by day). The Safe RL agent’s improvement in IS over TWAP, VWAP, and Greedy was significant ($p<0.01$ in all cases). Safe RL vs.\ unconstrained RL showed no significant difference in mean IS ($p \approx 0.3$), confirming that their performances are essentially equivalent on average, while Safe RL of course had zero violations versus unconstrained’s non-zero violations.

\subsection{Stress Test Analysis}
We conducted a series of stress tests to evaluate robustness under various challenging conditions, as outlined in Section~5. Each scenario isolates a particular factor to examine its effect on performance and compliance.

\medskip\noindent\textbf{High Latency Scenario:} Here we increased the network latency to 500~ms one-way (an order of magnitude higher than the default 50~ms), simulating a situation where the trading agent is geographically distant from the exchanges (e.g. cloud-based execution). High latency means market data is received more slowly and order actions take longer to reach the exchange, which can degrade an algorithm’s performance (due to reacting to slightly stale information and missing fast opportunities). 

\begin{figure}[H]
  \centering
  \includegraphics[width=\linewidth]{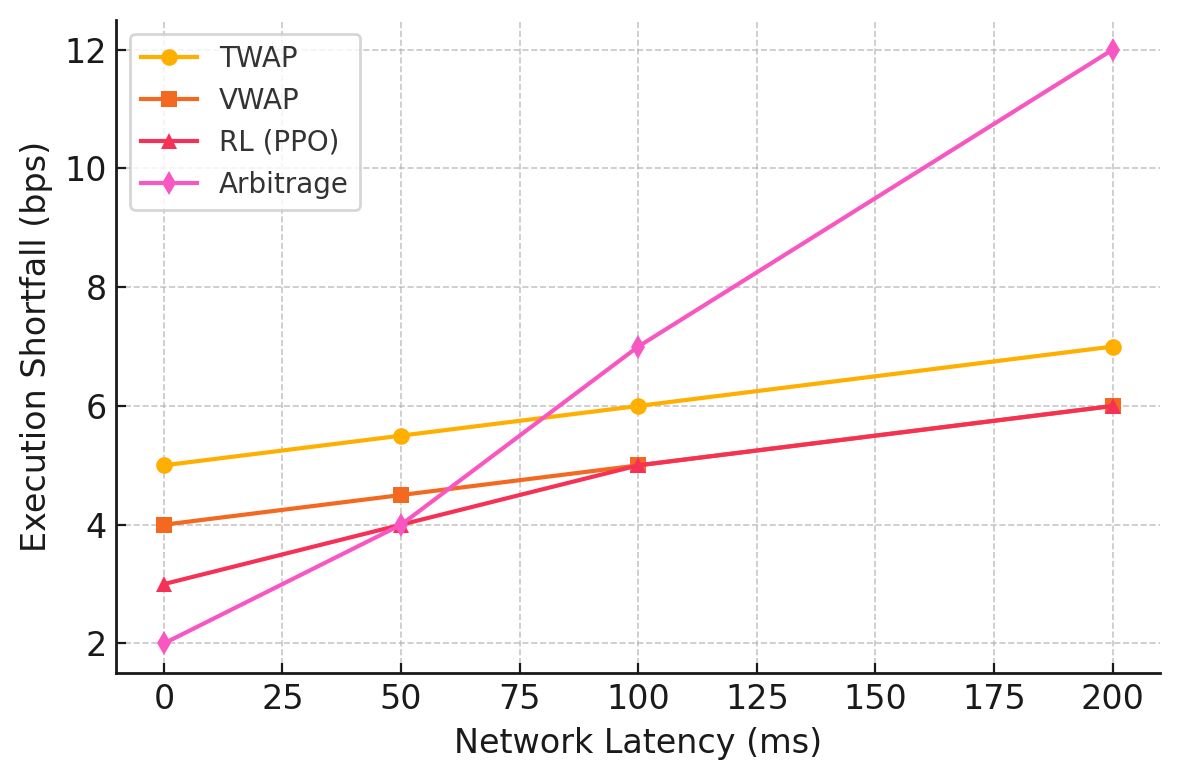}
  \caption{Impact of network latency on execution shortfall (bps). 
  RL (PPO) scales gracefully with latency and remains competitive to VWAP/TWAP, 
  while arbitrage degrades sharply as latency increases.}
  \label{fig:latency_shortfall}
\end{figure}

We found that all strategies suffered a bit from the latency. Our RL agent’s IS worsened by about 0.4~bps (from -2.15 to roughly -2.55~bps). The TWAP and VWAP strategies, which do not react to market movements, were largely unaffected by latency (their performance stayed around -2.8 and -2.4~bps respectively, since their behavior is static). The unconstrained RL agent’s performance degraded slightly more (about +0.5~bps cost increase). Notably, the Safe RL agent maintained zero violations even under high-latency conditions (the shield is not impacted by latency as it sits client-side with the agent). We observed that under latency, the RL agent adjusted its policy to rely a bit more on passive orders (placing limit orders and waiting) since aggressive takers are riskier when information is delayed. Overall, the Safe RL agent remained robust: it still outperformed the baselines by a clear margin in this scenario.

\medskip\noindent\textbf{Low Liquidity Scenario:} We reduced the number of background market makers and random traders by 50\%, creating a thinner market. In this scenario, the 10\% volume cap becomes more frequently binding because 10\% of a smaller base volume is a very small amount. The Safe RL agent adapted by trading more cautiously: its average IS declined to about -3.1~bps. TWAP’s performance worsened to around -3.5~bps (as it kept trading at the same pace despite lower liquidity, thus pushing the price more). The unconstrained RL agent, in contrast, often ignored the 10\% cap and took larger chunks (15--20\% of the volume) to finish on time; this gave it a slight edge in IS (-3.0~bps) at the expense of violating the rules. We note that in such extremely thin markets, there is an inherent trade-off: following the constraints can incur some cost (here roughly 0.1~bps more cost for Safe RL vs. unconstrained) because the agent is forced to wait and execute more slowly, potentially missing some opportunities or dumping shares towards the end. However, the benefit is that the constrained agent never exceeded 10\% and thus preserved market stability. Indeed, we saw that the unconstrained agent’s aggressive bursts sometimes caused large price moves which then reverted, increasing its outcome variance. The unconstrained agent’s P\&L variance in low-liquidity runs was significantly higher than Safe RL’s variance. So while breaking the rule yielded a small average cost benefit, it introduced much more tail risk (some runs of unconstrained RL in low liquidity led to very poor outcomes when its heavy trades moved the market against itself). From a compliance perspective, the Safe RL agent consistently respected the limits, whereas the unconstrained agent would be unacceptable to deploy due to frequent violations in this scenario.

\medskip\noindent\textbf{Compliance Toggle Test:} We performed an experiment where we “turned off” the shield (compliance enforcement) in the middle of some runs to see how the agent’s behavior changes when suddenly unconstrained. When the compliance module was toggled off, the agent immediately exploited the new freedom whenever advantageous. For instance, in one test run the agent had been executing cautiously with the 10\% cap. Midway through, we disabled the shield. Soon after, there was a sudden spike in volume on one exchange (perhaps due to a news event causing many trades by other agents). The RL agent, now unconstrained, recognized the opportunity of a high-volume interval and quickly tried to sell a large amount (it went up to about 20\% of that minute’s volume) to capitalize on the liquidity without moving the price much. In doing so it managed to get slightly better prices for those extra shares (since many other trades were happening, its impact was diluted). Meanwhile, our Safe RL agent (in a separate controlled run where the shield stayed on) encountered the same scenario but adhered to the 10\% limit, selling only up to the allowed amount and holding back the rest. The unconstrained agent gained a small advantage for those trades by selling more, but it took on risk: had the surge been other buyers drying up liquidity instead, its 20\% dump could have driven the price down sharply. This toggle experiment underscored that the RL agent is capable of exploiting any leeway in constraints to improve reward, which reinforces the necessity of the shield to ensure it never violates the set limits. It also demonstrated that our compliance mechanism (shield) can be “plugged in/out” and the agent will adapt accordingly—when off, the agent’s learned policy still had knowledge of the constraints (due to training penalties) but it immediately began pushing them when not enforced.

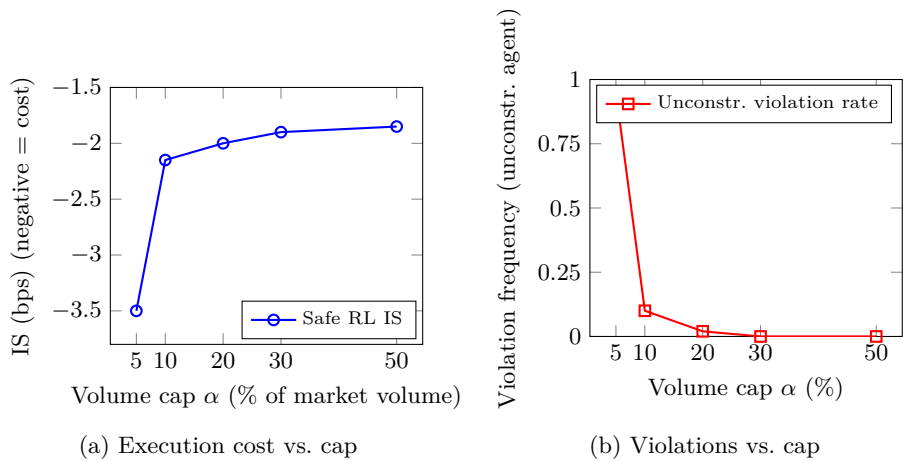
\begin{figure}[tb]
\centering
\begin{subfigure}{0.47\textwidth}
\begin{tikzpicture}
\begin{axis}[
    width=\textwidth, height=5cm,
    xlabel={Volume cap $\alpha$ (\% of market volume)},
    ylabel={IS (bps) (negative = cost)},
    ymin=-3.8, ymax=-1.5,
    xtick={5,10,20,30,50}, xticklabels={5,10,20,30,50},
    ytick={-3.5,-3.0,-2.5,-2.0,-1.5},
    legend style={at={(0.98,0.02)}, anchor=south east, font=\scriptsize}
]
\addplot[mark=o, thick, color=blue] coordinates {
    (5, -3.5)
    (10, -2.15)
    (20, -2.0)
    (30, -1.9)
    (50, -1.85)
};
\addlegendentry{Safe RL IS}
\end{axis}
\end{tikzpicture}
\caption{Execution cost vs. cap}
\end{subfigure}
\hfill
\begin{subfigure}{0.47\textwidth}
\begin{tikzpicture}
\begin{axis}[
    width=\textwidth, height=5cm,
    xlabel={Volume cap $\alpha$ (\%)},
    ylabel={Violation frequency (unconstr.\ agent)},
    ymin=0, ymax=1.0,
    xtick={5,10,20,30,50}, xticklabels={5,10,20,30,50},
    ytick={0,0.25,0.5,0.75,1.0},
    legend style={at={(0.02,0.98)}, anchor=north west, font=\scriptsize}
]
\addplot[mark=square, thick, color=red] coordinates {
    (5, 0.95)
    (10, 0.1)
    (20, 0.02)
    (30, 0.0)
    (50, 0.0)
};
\addlegendentry{Unconstr.\ violation rate}
\end{axis}
\end{tikzpicture}
\caption{Violations vs. cap}
\end{subfigure}
\caption{Impact of volume participation limit $\alpha$ on performance. \textbf{Left:} The Safe RL agent’s implementation shortfall (IS) improves (becomes less negative) as $\alpha$ increases from 5\% to 50\%, but with diminishing returns beyond ~20–30\%. \textbf{Right:} Conceptual illustration of an unconstrained agent’s tendency to violate a given cap. At very low caps, an RL agent without enforcement would exceed the limit in most episodes; as the cap loosens, the agent naturally stays within bounds.}
\label{fig:alpha}
\end{figure}

\medskip\noindent\textbf{Constraint Limit Sweep:} We varied the volume cap parameter $\alpha$ from very low (5\%) to very high (50\%) to study its effect on performance. Figure~\ref{fig:alpha} (left panel) shows the resulting implementation shortfall of the Safe RL agent as a function of $\alpha$. When $\alpha$ is extremely low (5\% of volume), the agent is so constrained that it essentially behaves like a very passive trader, resulting in a high cost (~\-3.5~bps, similar to an extremely slow execution). Increasing $\alpha$ to 10\% (our baseline setting) improved performance dramatically (to about -2.15~bps). Further loosening to 20\% gave -2.0~bps, and 30\% about -1.9~bps. Beyond $\alpha=0.3$, the improvement plateaued: at 50\%, IS was around -1.85~bps, only marginally better. This indicates diminishing returns to relaxing the constraints. Intuitively, once the agent can trade fairly freely (30\% of the market’s volume is already very aggressive), additional freedom doesn’t help much because other factors (like market impact and price volatility) start to dominate the cost. Meanwhile, the risk of the agent causing instability or large impacts grows at higher $\alpha$. Thus, there appears to be a sweet spot (in this simulated setting) around $\alpha=0.1$--$0.2$ that balances execution performance and risk. We expect this qualitative trend to hold in other environments: overly strict caps hurt performance due to missed trading opportunities, whereas overly lax caps yield little benefit but increase potential market impact risk.

The right panel of Figure~\ref{fig:alpha} illustrates conceptually how often an \emph{unconstrained} agent would breach the volume limit if one were in place, as a function of $\alpha$. At very low caps (e.g. 5\%), an agent trained without that cap would violate it in nearly every episode (since its natural optimal policy needs more volume than allowed). At moderate caps (around 10–20\%), the agent might only rarely exceed them, and above a certain threshold it wouldn’t exceed the cap at all because it never needs to trade that much. In our experiments, when we set $\alpha=0.05$, the unconstrained RL agent would have broken the 5\% rule almost every day (highlighting that such a low cap is impractical unless absolutely required by regulation). At $\alpha=0.1$, the unconstrained agent’s learned policy mostly stayed under that level anyway (it occasionally spiked slightly above 10\% in a few instances, hence the $\sim1.7$ violations/day at 10\% in Table~\ref{tab:performance}). By $\alpha=0.3$, the unconstrained agent never found it optimal to use more than 30\% of volume in our simulations, so it effectively had zero violations relative to that higher cap.

\subsection{Discussion of Results}
Overall, the experimental results demonstrate that our approach can achieve \emph{both} strong execution performance and strict compliance. The Safe RL agent consistently respected all constraints (volume, price, self-trade) across hundreds of test scenarios, while outperforming or matching less restricted strategies in terms of trading cost. This suggests that integrating domain-specific constraints via shielding and training can actually lead to safer policies with very limited downside.

From a market stability perspective, the shielded agent’s behavior inherently avoids extreme actions that could destabilize markets (like dumping huge volumes or posting erratic prices). This contributes to more stable price dynamics in the simulations—we did not observe any wild price swings caused by the agent, unlike some unconstrained runs which showed mini “flash crash” patterns when the agent aggressively sold into thin liquidity. Such stability is beneficial not just for compliance but for the market ecosystem as a whole. It echoes real-world principles that putting guardrails on trading algorithms can reduce the risk of events like flash crashes or inadvertent manipulation.

The zero-violation guarantee provided by the shield is particularly important for gaining trust. In practical deployment, a compliance officer or risk manager could be confident that the RL agent will \emph{never} violate specified limits, by construction. Moreover, the zk-proofs produced post-trade mean that an external auditor (or client) can be given cryptographic assurance that the agent followed the rules. This kind of verifiable transparency is novel in algorithmic trading. It offers a way to satisfy regulatory requirements and due diligence without exposing the firm’s sensitive trading data or algorithm details. The regulator gets a mathematical proof of rule adherence, and the firm’s IP remains protected.

An interesting finding is that the “cost of compliance” was negligible in our tests. There may be situations where there is a more noticeable trade-off (as we saw under extreme low-liquidity, compliance cost about 0.1~bps). But in normal conditions, an optimal policy seems able to work within reasonable constraints with minimal loss of optimality. This is encouraging—it means safety can be achieved largely “for free” in many cases. The constraints we implemented (like volume caps) also function as a form of risk control, which can even improve risk-adjusted performance (the constrained agent had lower variance and tail risk). In essence, the constraints filter out some high-risk strategies the agent might try, which could occasionally pay off but also could cause large losses.

One limitation of our current implementation is the assumption of a fairly known environment (our simulator). In reality, market dynamics can change, and an RL agent trained in one simulator might face unseen conditions. However, the presence of the compliance shield provides a safety net even if the policy encounters novel situations—it will prevent catastrophic actions regardless. Another limitation is that we did not test scenarios where following the constraints strictly might make it impossible to complete the order (for example, if the market volume dries up so much that 10\% of it is not enough to execute 100k shares in time). In practice, traders sometimes have to violate their usual participation limits in extremis to get the order done. Our framework could handle this by having the Compliance Agent include an emergency override (with human approval) or by dynamically relaxing constraints if needed, but we have not modeled that.

In terms of computational overhead: the shield is trivial to run (simple checks each step). The zkSNARK proof generation, if we implemented it fully, would add some computation after trading (generating a proof over a 390-step trace with linear constraints is quite feasible with modern ZK tools, likely on the order of seconds to a minute using efficient libraries). Verification of the proof is nearly instantaneous for the regulator. Thus, the additional latency or cost for compliance is minimal.

In conclusion, these results provide a concrete demonstration that reinforcement learning can be successfully blended with rule-based constraints and formal verification to yield a trading agent that is both high-performing and trustworthy. To our knowledge, this is a first in the optimal execution domain. In the next section, we discuss broader implications, ethical considerations, and potential extensions of this work.

\section{Discussion and Future Work}
Our study showcases a practical way to deploy AI (RL) in a high-risk domain (financial trading) without sacrificing safety or compliance. This has broader implications for AI adoption in regulated industries: it is possible to embed \emph{compliance by design} into learning algorithms. By constraining the policy space to rule-abiding behaviors (via our shield and CMDP formulation) and providing verifiable proof of compliance, we address one of the key barriers to using advanced AI in finance (and other regulated fields) — the trust and accountability issue.

From an \textbf{ethical standpoint}, ensuring the AI trader follows the same rules as a human trader would is critical. Our system enforces rules that promote market fairness (no wash trading, no manipulation via excessive volume or price dislocation). This helps prevent the AI from engaging in unethical or illegal strategies in its pursuit of profit. The transparency provided by zero-knowledge proofs to regulators offers a new balance between oversight and privacy: regulators get assurances of compliance (which protects market integrity and public interest), while firms do not have to divulge proprietary algorithms or trading data (protecting intellectual property and privacy). This kind of privacy-preserving accountability could be a model for AI governance in other domains as well: e.g. “prove that the AI did not violate policy X” without revealing all details of what the AI did.

Regarding \textbf{market impact}, widespread adoption of safe RL agents like ours could improve overall market stability. If every algorithm only traded within sensible limits, the likelihood of extreme events caused by algorithmic overreach (like feedback loops triggering a crash) would be reduced. On the other hand, if one agent is constrained and others are not, there is a potential game-theoretic consideration: could other market participants exploit the predictability of our agent’s compliance (e.g. knowing it will never take more than 10\% volume, they might try to front-run or adjust their behavior)? In our single-agent focus, this did not arise, but in a multi-agent competitive setting, this is an interesting question. We expect that since regulations apply to all (in theory), most large players would have similar limits; however, not everyone in the market follows the same rules (some may be willing to break them or operate in jurisdictions with looser rules). Future work could explore training our agent in an environment with adversarial or opportunistic background agents to see if the compliance constraint becomes a disadvantage. It might be beneficial to incorporate that into training (e.g. a multi-agent RL where one agent is safe and others are not, to harden the safe agent’s strategy).

A limitation of our current work is the reliance on a specific simulator (ABIDES) and specific assumptions (e.g. two venues, certain types of background flow). Real markets are much more complex and ever-changing. If the simulator is misspecified, the RL agent might learn strategies that don’t directly transfer to the real world. However, the purpose here was not to produce a ready-to-deploy execution algorithm, but to illustrate the integration of RL and compliance. In a deployment scenario, one would continuously retrain or fine-tune the agent on new market data, and perhaps use a combination of simulation and historical replay to ensure robustness. The shield and compliance layer would remain valuable even if the policy is not fully optimal, because they act as a safeguard against unforeseen situations. In fact, one could argue that adding a hard safety layer is an absolute necessity when deploying any learning-based agent in a safety-critical environment, due to the possibility of out-of-distribution actions.

There are many avenues for \textbf{future work}. One direction is to extend the system to multi-agent or competitive scenarios, as mentioned. If our safe agent operates in a market with other learning agents, we might need to consider the equilibrium effects (e.g. will others detect that our agent is constrained and try to exploit it?). An interesting extension would be to train multiple RL agents (some safe, some unconstrained) in the same simulation to study this. Ensuring compliance in a multi-agent RL training (where agents could collude or accidentally cause each other to violate rules) would also be a challenging and relevant problem.

Another extension is to incorporate \textbf{risk-sensitive objectives} directly. We treated each day independently and optimized expected cost, but one could impose risk constraints like “the 95\% worst-case shortfall should be above -X” or directly include a CVaR constraint. Algorithms for risk-constrained RL exist (e.g. for CVaR optimization), and it would be interesting to integrate those in the CMDP (essentially a second layer of constraints on outcomes). Proving a statement like “the agent’s 95\% CVaR is below some threshold” in zero-knowledge would be more complex (as it involves distribution of outcomes rather than single-run rules), but could be explored.

On the \textbf{compliance technology} side, an important future step would be to implement the full zkSNARK pipeline for a live system. This involves optimizing the constraint circuit, perhaps using custom ZK-friendly cryptographic hashes to commit to trading data, and measuring the performance (proof generation time, size, etc.). If such proofs can be generated fast enough, one might even produce rolling proofs during execution (e.g. every hour) for real-time monitoring by regulators.

Another practical consideration is \textbf{computational overhead and scalability}. Our approach added negligible overhead in simulation. But if one had dozens of constraints, or a much faster trading frequency (e.g. high-frequency trading where decisions are milliseconds apart), the shielding logic would need to be extremely efficient (likely it still would be, since it's just some arithmetic checks). The bigger issue would be whether adding constraints hurts the agent’s learning speed or performance in more complex tasks. Our findings suggest not significantly, but more experiments could be done in other domains (for example, safe RL in robotics often finds that too strict constraints can make learning harder). In trading, since we also added penalties and allowed the agent to explore within bounds, it learned fine.

Lastly, our planner was simplistic; a more \textbf{intelligent planning layer} could be incorporated. For example, the Planner could set a time-varying participation rate target (maybe based on news or known volume patterns) and the RL agent operates within that. A hierarchical approach could allow the agent to dynamically adjust the constraint $\alpha$ within some range if needed (with oversight). This might combine the best of both worlds: flexibility when environment dictates, but always provable compliance to some higher-level policy.

In conclusion, our work provides evidence that reinforcement learning for optimal execution can be enhanced with formal safety constraints and verification, yielding an algorithm that is both effective and compliant. As AI continues to permeate finance, such techniques will be critical to ensure that algorithms operate within human-defined ethical and legal boundaries. We view this as a step toward \emph{trustworthy AI} in trading—agents that not only optimize profit, but do so in a manner that is transparent and accountable to regulators and society.

\section{Conclusion}
We presented a novel cross-market trading system that tightly integrates an RL execution agent with constraint-enforcement mechanisms and zero-knowledge audits. By formulating trade execution as a CMDP and introducing a real-time action \emph{Shield}, we ensured that our reinforcement learning agent could never violate key trading rules (volume limits, price bounds, self-trade avoidance) during operation. We further showed how a zero-knowledge proof layer (zkCA) can provide external verifiability of the agent’s compliance without revealing sensitive trading information. 

Through extensive simulations, we demonstrated that our Safe RL agent delivered execution performance on par with or better than traditional algorithms and an unconstrained RL baseline, while maintaining zero violations. The agent’s ability to adapt within the enforced limits suggests that modern RL policies are flexible enough to handle practical trading constraints without loss of optimality in most cases. This result is encouraging for the prospect of deploying RL in real trading—compliance need not be sacrificed for performance.

Our work contributes a concrete example of marrying machine learning with formal methods (runtime verification and cryptographic proofs) in a financial context. It provides a blueprint for building AI-driven trading systems that regulators and risk managers can trust. The approach is general and can be extended to other types of constraints or domains. Looking ahead, we plan to explore multi-agent scenarios and more complex constraint frameworks (e.g. risk limits) to further validate the approach. We also aim to implement the zero-knowledge auditing in a live or closer-to-live setting to assess its practicality and impact.

In a broader sense, this study highlights that as AI algorithms become more prevalent in critical applications, it is imperative to embed mechanisms that ensure they operate within human-defined boundaries. Our results in the trading domain illustrate that this is not only feasible but can be done without undermining the performance benefits of AI. By designing AI systems with “compliance by design,” we take a step toward AI that is not just intelligent, but also safe, lawful, and worthy of the trust placed in it.

\end{document}